\documentclass[10pt,conference]{IEEEtran}
\IEEEoverridecommandlockouts
\usepackage{cite}
\usepackage{amsmath,amssymb,amsfonts}
\usepackage{algorithmic,algorithm}
\usepackage{graphicx}
\usepackage{textcomp}
\usepackage{xcolor}
\usepackage{hyperref}
\usepackage{url}
\usepackage{listings}
\usepackage{fancyvrb}
\usepackage{booktabs}

\usepackage[subrefformat=parens]{subcaption}
\usepackage{amsthm}
\theoremstyle{plain}

\newtheorem{dfn}{Definition}[section]

\def\Nonterm{\mathcal{N}}
\newcommand{\defeq}{\mathrel{:\mkern-0.25mu=}}

\title{TruncProof: A Guardrail for LLM-based JSON Generation under Token-Length Constraints}

\begin{document}

\author{\IEEEauthorblockN{Yoshio Kato and Shuhei Tarashima}
\IEEEauthorblockA{NTT DOCOMO BUSINESS, Inc., Japan\\
{\tt\small yoshio.kato@ntt.com, tarashima@acm.org}
}}

\maketitle

\begin{abstract}
The LLM-based generation of machine-readable outputs such as JSON has attracted significant attention for integration with external systems. 
However, existing approaches cannot strictly enforce the maximum number of tokens to be generated, leading to infinite generation or truncated outputs that cause a system malfunction.
To address this limitation, we propose TruncProof, a novel grammar-constrained generation method that enables LLMs to produce grammatically valid JSONs while adhering to a predefined token limit.
By leveraging the properties of LL(1) parsers, TruncProof efficiently approximates the minimum number of tokens required to complete a grammatically valid output at each decoding step.
Experiments on the Text-to-JSON instruction tasks
demonstrate that TruncProof successfully generates syntactically correct outputs even under strict token constraints.
Furthermore, we show that TruncProof can be effectively combined with advanced decoding strategies, resulting in outputs that are not only grammatically valid but also semantically accurate.
The source code is public at \url{https://github.com/Yosshi999/TruncProof}.
\end{abstract}


\section{Introduction}
\label{sec:intro}
Recently, there has been a growing body of research on solving complex tasks by combining the text generation capabilities of large language models (LLMs) with external tools \cite{MathCoder,VisProg}.
In these use-cases, LLMs are expected to consistently produce well-formed, machine-readable outputs in accordance with specified grammars.
Among these formats, JSON is the de facto standard, widely supported by various providers \cite{openai_jsonmode, claude_jsonmode, gemini_jsonmode} and commonly used as a messaging protocol between LLM-based agents and external systems \cite{mcp}.
However, because LLM tokenizers are designed for natural language, it is difficult to reliably enforce grammatically valid JSON output through fine-tuning or prompting alone.
To address this robustness issue, several grammar-constrained generation (GCG) guardrails for JSON have been proposed in the literature \cite{guidance,Outlines,llamacpp,domino,SynCode,XGrammar,llguidance}.
\par
While these methods can enforce complex grammatical constraints, they all have a critical limitation: {\it they cannot strictly enforce a maximum number of generated tokens}.
In practical applications, imposing a token limit is essential to prevent infinite generation, control memory usage, and keep the output within the model’s context window.
However, because current constraint-based methods cannot dynamically estimate the number of tokens needed to complete a grammatically valid output, they terminate generation abruptly once the token limit is reached, often resulting in incomplete or grammatically invalid outputs.
This issue is particularly problematic in agent-based applications, where autonomous agents are required to quickly exchange JSON without human intervention; such termination leads to parse errors that can subsequently disrupt downstream processes.
\par
To address this truncation issue, we propose a novel GCG guardrail that enables LLMs to generate grammatically correct JSONs while adhering to a specified maximum number of tokens.
This requires estimating, at each decoding step, the minimum number of tokens needed to complete a grammatically valid output.
We address this challenge by leveraging the fact that a JSON parser can be implemented as LL(1) \cite{parsetheory}.
LL(1) is simpler than the parsing strategies employed in existing parsers ({\it e.g.}, LALR(1)\cite{lalr1} is used in \cite{Lark}) and can therefore efficiently compute the shortest valid token sequence required to complete the output at each step, and construct constraint masks to prevent the selection of tokens that would violate the grammar or token limit.
\par
Our proposed method, called TruncProof hereafter, has a form of logit modifier.
Therefore, it is compatible with a wide range of tokenizers, language models, other logit modifiers and various decoding strategies.
We evaluate TruncProof on the Text-to-JSON instruction tasks \cite{NousResearch}, and
experimental results show that TruncProof enables LLMs ({\it e.g.}, Gemma2\cite{gemma_2024}, Llama2\cite{llama2}) to produce grammatically valid JSON outputs, even under strict token budget constraints, whereas existing methods almost fail to do so.
We also demonstrate that TruncProof can be integrated into advanced decoding strategies such as Beam Search and Monte Carlo Tree Search, significantly enhancing the semantic robustness of the generated JSON while preserving its grammatical validity.
\section{Background}
\label{sec:background}
To enhance self-containment, we first introduce the foundation of Grammar-Constrained Generation (GCG) in \S\ref{sec:background:constr}. 
We then provide an overview of Context-Free Grammars (CFG) in \S\ref{sec:background:cfg}, followed by implementations of its parsers in \S\ref{sec:background:cfg-imple}, to formally define the LL(1) parser employed in our TruncProof.
Throughout this paper, we denote the finite set of characters that can be generated by an LLM as \begin{math}\Sigma\end{math}, and the set of all finite-length strings over \begin{math}\Sigma\end{math} as \begin{math}\Sigma^*\end{math}
\footnote{
For example, when \begin{math}\Sigma=\{\mbox{a},\mbox{b},\mbox{c}\}\end{math}, \begin{math}\Sigma^*=\{\epsilon, \mbox{a}, \mbox{b}, \mbox{c}, \mbox{aa}, \mbox{ab}, \mbox{ac}, \mbox{ba}, \cdots \}\end{math}.
}.
The empty string is denoted by \begin{math}\epsilon\end{math}, and the concatenation of two strings $w,v$ is represented as \begin{math}(w.v)\end{math}.

\subsection{Grammar-Constrained Generation (GCG)} 
\label{sec:background:constr}
Modern LLMs generate output tokens from a vocabulary $\mathcal{V}$ in an auto-regressive manner:
At each generation step $i$, the model takes the current partial output \begin{math}t_{<i} = t_1. \cdots .t_{i-1}\in\mathcal{V}^*\end{math} and predicts the probability distribution of the $i$-th token \begin{math}P(t_i \mid t_{<i})\end{math}.
In GCG,
\textit{constraint functions} evaluate the grammatical validity of each candidate token $t_i$ at every step.
Specifically, given a string $t_{<i}$, the constraint function uses a \textit{parser} to check whether there exists a string $w$ that extends the candidate token into a grammatically valid sentence, and returns the result in the form of a \textit{constraint mask} ${\bf m}$.
Formally, the element of ${\bf m}$ for a next token candidate $t$, $m_t$, is defined as follows:
\begin{equation} \label{eqn:grammarmask}
m_t = true \ \Rightarrow \  \exists w \in \mathcal{V}^* \ \mbox{s.t.}\ (t_{<i} . t . w)\in L(G),
\end{equation}
where $G$ is a grammar and $L(G)$ is the \textit{language} defined as the set of strings accepted by $G$.
Tokens deemed grammatically invalid are re-assigned zero probability by element-wise multiplication between the probability distribution and the constraint mask {\it i.e.}, $P(t_i \mid t_{<i}) \odot {\bf m}$.
Note that this modification is applied prior to selecting the next token for generation. 
Consequently,
from an algorithmic perspective,
any GCG method, including our proposed TruncProof, can be combined with various decoding strategies. 
\subsection{Context-Free Grammar (CFG)}
\label{sec:background:cfg}
CFG has been used to define a variety of machine-readable formats.
CFG is characterized by a four-tuple \begin{math}(\Nonterm, \Sigma_T, R, S)\end{math}:
a finite set of the \textit{nonterminal} symbols that does not appear in the language $\Nonterm$,
a finite set of the \textit{terminal} symbols as the alphabet in the language $\Sigma_T$,
a finite relation which represents derivation rules that rewrite a single nonterminal to the terminal or nonterminal symbols with 0 or more length $R\subset \Nonterm \times (\Nonterm\cup\Sigma_T)^*$, 
and the start symbol $S \in \Nonterm$.
Using this expression, we can define the language $L(G)$ as the set of the terminal sequences.
Any terminal sequence $\sigma\in\Sigma_T^*$ in the language can be generated by repeated derivations (denoted as $\rightarrow^*$) from the start symbol.
CFG parsers must construct a derivation process that generates the string from the start symbol to determine whether the string belongs to the language.
Notice that these processes can be visualized as derivation trees, with the start symbol at the root and terminal symbols at the leaves.
\par
Usually, to prevent grammars being too complicated, terminal symbols in CFG are defined as \textit{Regular Expression (Regex)}
instead of characters \cite{Lark} and the parsers preprocess the input string to identify the equivalent terminal sequence.
Regex can be parsed by using 
\textit{Deterministic Finite Automaton (DFA)}, 
which characterized by a five-tuple \begin{math}(Q, \Sigma, \delta, q_0, F)\end{math}: 
a finite set of states $Q$, a finite set of recognizable characters $\Sigma$, a transition function that determines the next state based on a current state and a captured character $\delta: Q\times \Sigma \rightarrow Q$, the initial state $q_0 \in Q$, and a set of accepting states $F\subseteq Q$.
DFA starts from the initial state and accepts the input if and only if its state transitions to an accepting state by processing each character one by one.

\subsection{Implementations of CFG parsers}
\label{sec:background:cfg-imple}
There are two primary approaches to implement CFG parsers \cite{parsetheory}:
\textbf{The bottom-up approach}, such as LALR(1) parsers, which identifies the derivation tree from the bottom ({\it i.e.}, from the leaf nodes), and \textbf{the top-down approach}, such as LL(1) parsers, which constructs the derivation tree from its top ({\it i.e.}, from the root).
Contrary to bottom-up parsers, top-down parsers can easily enumerate possible continuations of the current input by applying arbitrary derivations from the unexpanded nonterminals.
Therefore, to leverage this advantage, in our TruncProof we employ LL(1), a top-down parser that permits only single-terminal lookahead without allowing backtracking ({\it i.e.}, reconstruction of the derivation tree).
Although LL(1) does not support all Context-Free languages, it is still sufficiently expressive for deeply nested structures such as JSON.
A formal definition of LL(1) parser is as follows.
\begin{dfn}[LL(1) parser]
A LL(1) parser that recognize a CFG $(\Nonterm, \Sigma_T, R, S)$ has an input buffer and a stack $\Gamma \in (\Nonterm \cup \Sigma_T)^*$ initialized with $(S)$. For each step, the parser reads a terminal $x$ from input and selects one of the following operations.
\begin{itemize}
    \item If the top of stack is a terminal, it must be same as input $x$, and pop it from the stack.
    \item If the top of stack is a nonterminal $A$, there must be a unique sequence of derivations $A\rightarrow^* x\beta$, and replace $A$ with the symbols that make up $\beta$ on the stack.
    \item Otherwise the parsing is failed.
\end{itemize}
If both input and stack get empty, it has parsed the input successfully.
\end{dfn}

\begin{figure*}[h]
\centering
\includegraphics[width=\linewidth]{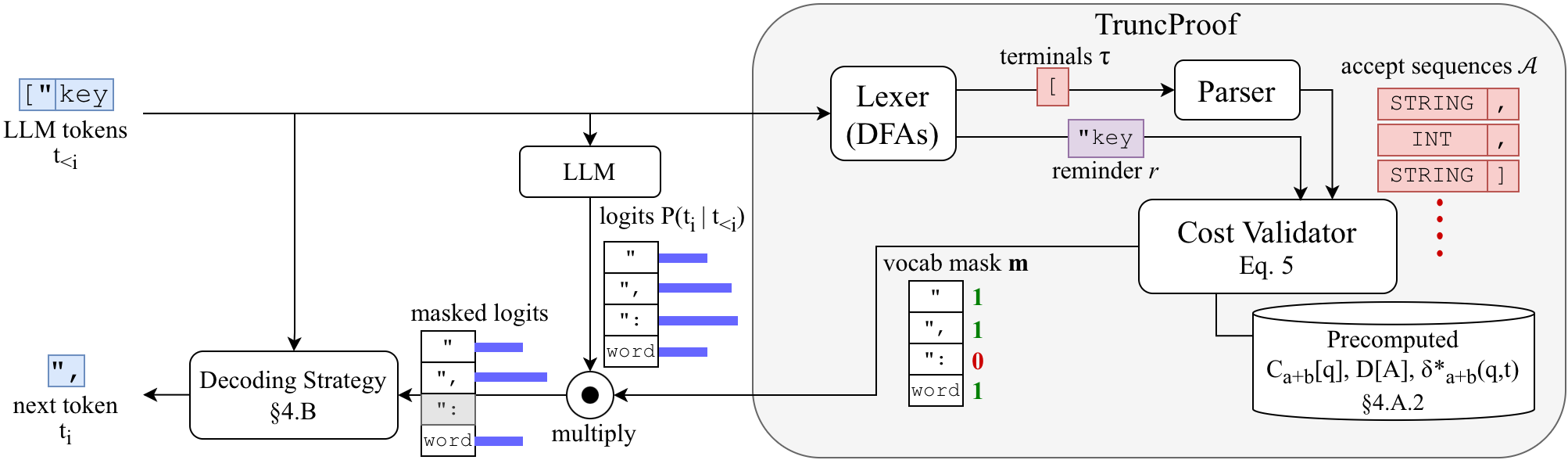}
\caption{Overview of TruncProof. For $i$-th generation step, Lexer parses the intermediate LLM tokens generated by the LLM into the terminals $\tau$ and the remainder $r$, Parser collects all possible terminal sequences (called accept sequences $\mathcal{A}$) whose length is at most two, and the Cost Validator constructs the vocabulary mask $\mathbf{m}$ by validating the future cost for each candidate token based on the precomputed cache.}
\label{fig:diagram}
\end{figure*}

\section{Related Works}
\label{sec:related}
Several GCG methods have been proposed in recent years, most of which can be classified based on the type of grammar they support.
For example, PICARD \cite{picard} is designed for SQL, where it generates multiple candidates simultaneously and checks their parsability.
LMQL \cite{lmql} allows user-defined grammars based on Regex through a custom specification language. 
Outlines \cite{Outlines} improves the efficiency of Regex-based generation by precomputing valid token sets for each DFA state.
Although Outlines also supports CFGs, it is usually slow since it repeats sampling and validation of candidates until a grammatically valid token is found.
Recently, research has been conducted to further optimize precomputation or runtime processing within the scope of CFGs:
DOMINO \cite{domino} and SynCode \cite{SynCode} integrate optimized Regex validation with the CFG parsers that enumerate acceptable terminal sequences.
XGrammar \cite{XGrammar} introduces a variant of CFG parser that operates on characters rather than terminals, thereby reducing the overhead associated with terminal processing.
LLGuidance \cite{llguidance} adopts trie trees to handle LLM tokens with low-level optimization to reduce the overhead in runtime.
GreatGramma \cite{park2025flexible} aggregates all terminal definitions and the LLM vocabulary into a single Finit State Transducer that processes input token by token, which largely reduces the preprocessing cost.
\par
While the above methods can impose sufficiently complex grammatical constraints on LLMs, they share a common limitation: they cannot ensure that generation halts within a specified number of tokens. 
Notice that IterGen \cite{itergen} addresses this problem by repeatedly regenerating outputs until a desired result is obtained. 
However, it does not guarantee that a grammatically correct output will be found within a reasonable number of iterations.
\par
We also note that the literature includes methods that extend beyond CFG-based constraints.
\cite{mundler2025typeaware} and \cite{li2025correctnessguaranteed} propose a code generation framework that imposes richer constraints than CFGs, aiming to avoid any errors during compilation or execution.
While this direction is promising, these methods abandon constraint mask generation and instead rely on inefficient candidate sampling, similar to Outlines, which is especially disadvantageous when combined with advanced decoding strategies.
\cite{geng2023grammarconstrained} introduces token-level grammars that directly provide next valid tokens and supports more flexible grammars than CFGs.
However, this token-level approach potentially results in worse perplexity, since it prohibits to generate the same string consisting of natural token combinations.
\section{TruncProof}
\label{sec:proposed}

Let a grammar $G$ be specified in the form of an LL(1) grammar $(\Nonterm, \Sigma_T, R, S)$.
We assume that each terminal symbol in $\Sigma_T$ is defined by a Regex; 
For each terminal, there exists a corresponding DFA $\mathcal{M}_a\defeq(Q_a, \Sigma, \delta_a, q_{a0}, F_a)$ that accepts the strings defined by the Regex.
Given a grammatically valid partial output \begin{math}t_{<i}\end{math}, our TruncProof serves as a constraint function that returns the binary mask ${\bf m}$,
where each entry $m_t$ represents the grammatical validity of a token $t\in\mathcal{V}$ within the pre-defined token limit $N_{max}$.
By extending Equation~\ref{eqn:grammarmask}, $m_t$ is formally defined as follows:
\begin{equation}\begin{array}{l}
m_t = true \ \Rightarrow \\
\exists w \in \mathcal{V}^* \ \mbox{s.t.}\ \left((t_{<i} . t . w)\in L(G)\ \mbox{and} \ |t_{<i} . t . w| \leq N_{max}\right). \label{eqn:mask}
\end{array}\end{equation}
This mask can be used to filter out tokens that would result in either (1) a grammatically invalid continuation or (2) an output exceeding $N_{max}$.
\par
In \S~\ref{sec:proposed:limit}, we describe the details of TruncProof, which returns the mask ${\bf m}$. 
Note that this mask ensures grammatical validity but does not fully account for semantic correctness. 
To produce outputs that are both grammatically valid and semantically coherent, we extend TruncProof with advanced decoding strategies, which is detailed in \S~\ref{sec:strategy}.

\subsection{Details of TruncProof} 
\label{sec:proposed:limit}

\begin{figure*}[h]
    \centering
    \vspace{-0.4cm}
    \begin{minipage}[t]{\columnwidth} 
    \centering
    \includegraphics[width=0.9\columnwidth]{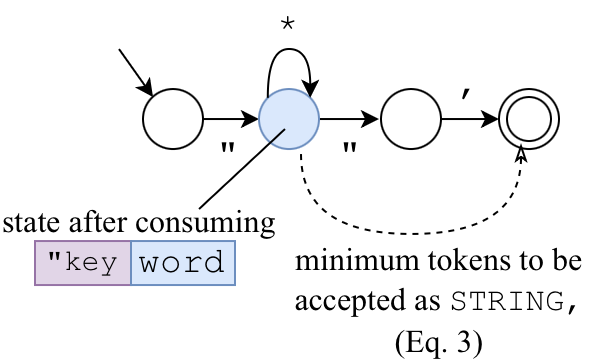}
    \subcaption{DFA state after accepting \fbox{``keyword}}
    \label{fig:sketch-a}
    \end{minipage}
    \begin{minipage}[t]{\columnwidth} 
    \centering
    \includegraphics[width=0.9\columnwidth]{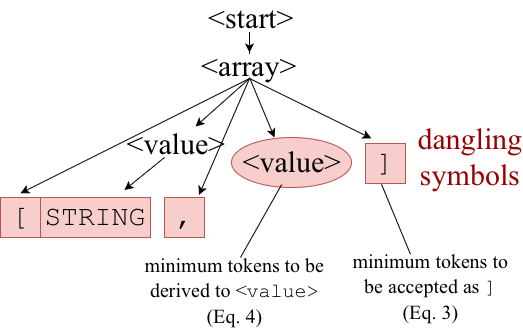}
    \subcaption{Parser state after accepting \fbox{[``keyword",}}
    \label{fig:sketch-b}
    \end{minipage}
    \caption{The examples of counting the future tokens in Cost Validator illustrated in Figure~\ref{fig:diagram}.}
    \label{fig:jsontree}
\end{figure*}

Figure~\ref{fig:diagram} illustrates the overall structure of TruncProof.
In runtime, the following steps are executed iteratively within the generation loop: 
(i) Given the intermediate output generated by the LLM, Lexer that handles Regex and Parser that handles LL(1) grammar incrementally parse the newly generated token based on the terminal sequence obtained in the previous iteration.
(ii) A Cost Validator estimates the number of tokens needed in the future assuming a next token (as illustrated in Figure~\ref{fig:jsontree}), and
verifies whether the generated output remains grammatically valid under the specified token budget.

To efficiently operate the Cost Validator, we precompute the estimation of the shortest token lengths for realizing any terminal and nonterminal defined by the given LL(1) grammar.
In the following sections we describe the behavior in the runtime phase and the things to be prepared in the precomputation phase.

\subsubsection{Runtime Phase}
As shown in Figure~\ref{fig:diagram}, we first divide the intermediate input $t_{<i}$ into the terminal sequence $\tau\in \Sigma_T^*$ and the 
remainder\footnote{
User-defined terminal symbols may not align exactly with LLM tokens. 
In such cases, some suffixes of the output remain unprocessed as remainders.
} $r\in \Sigma^*$ by using the DFAs, then partially parse $\tau$ to identify the derivation tree by using the LL(1) parser.
This process can be executed incrementally by using the results in the previous iteration.
Next, we enumerate the terminal sequences with a length of at most two {\it i.e.}, $a, b\in \Sigma_T$, that can be given to the current parser in this generation step.
The reason why we take two-length terminals in consideration is because
this extension allows us to better exploit the generative capabilities of the LLM while the relaxed constraint still ensures the condition defined in Equation~\ref{eqn:mask}.
We hereafter call the set of the sequences as \textit{accept sequence} $\mathcal{A} \subseteq \Sigma_T \cup\Sigma_T^2$.
After that, we calculate the two types of cost to complete the generation: the number of tokens to complete the remainder as terminals $(a,b)$ (as illustrated in Figure~\ref{fig:sketch-a}), and the further cost $d_{cost}(\tau. a. b)$ to complete the whole string after $a$ and $b$ are accepted by the parser (as illustrated in Figure~\ref{fig:sketch-b}).
The former cost can be estimated as the minimum number of tokens required to transition from each state $q$ in the corresponding DFA $M_{a+b}$ to an accepting state, which is formulated as follows:
\begin{equation}\label{eqn:c}
    C_{a+b}[q] \defeq \left\{\begin{array}{l}
        \min_{w \in \mathcal{V}^*} |w| \ \  \mbox{subject\ to\ } \delta_{a+b}^*(q, w) \in F_{a+b}  \\
        \multicolumn{1}{r}{(\mbox{if\ } \exists w \ \mbox{s.t.} \ \delta_{a+b}^*(q, w) \in F_{a+b})} \\
        \infty \ \ (\mbox{otherwise}),
    \end{array}\right.
\end{equation}
where $\delta^{*}_{a}$ is an iterated transition function {\it i.e.},
\begin{math}\delta^{*}_{a}(q, x_1.\cdots .x_n) = \delta_{a}(\cdots\delta_{a}(q, x_1)\cdots,x_n) \end{math}.
If there is no token sequence $w$ which can reach to any accepting state from $q$, $C_{a+b}[q]$ is set to infinity.
This ensures that grammatically invalid tokens are automatically excluded due to their infinity cost.
The latter cost $d_{cost}(\tau. a. b)$ is computed as the sum of the minimum number of tokens to consume the terminals and nonterminals that remains unresolved by the LL(1) parser (the dangling symbols illustrated in Figure~\ref{fig:sketch-b}).
To compute it, we need the approximate shortest token length $D[A]$ derivable from each nonterminal $A\in\mathcal{N}$, by the following equation:
\begin{equation}\label{eqn:d}
    D[A] \defeq \min_{\sigma \in \Sigma_T^*} \sum_{i=1}^{|\sigma|} C_{\sigma_i}[q_{\sigma_i0}] \ \ \mbox{subject\ to\ } A \rightarrow^* \sigma,
\end{equation}
where $\sigma_i$ denotes the $i$-th terminal symbol in the sequence $\sigma$.
In summary, the entry of the constraint mask ${\bf m}^{(a,b)}$ for a token $t$, {\it i.e.}, $m^{(a,b)}_t$, is computed as follows:
\begin{equation}\label{eqn:fullcost}
    \begin{split}
    &m^{(a,b)}_t \defeq true \ \mbox{iff.}  \\
    &\ \begin{array}{cc}
    i & \texttt{(consumed tokens)}\\
    + C_{a+b}[\delta_{a+b}^*(q_{a+b0}, r.t)] & \texttt{(tokens DFA accepts)}\\
    + d_{cost}(\tau. a. b) & \texttt{(tokens to terminate)} \\
    & < N_{max}, \\
    \end{array}\\
    \end{split}
\end{equation}
where $i$ is the number of generated tokens.
Once the simulation of the parser and the calculation of the future cost are performed, the constraint mask ${\bf m}$ can be obtained by taking the element-wise union of the masks ${\bf m}^{(a,b)}$ for each $(a,b)\in\mathcal{A}$.
Since each valid entry corresponds an actual sequence of tokens, it guarantees the result that adheres to the grammar and token limit.

\textbf{Time Complexity Analysis.}
At each iteration of the generation loop, the computational bottleneck is the simulation of the LL(1) parser to calculate $d_{cost}(\tau.a.b)$ for each $(a,b)\in\mathcal{A}$.
It takes $O(|\Sigma_T|^2(T_G + |\Gamma|))$, where $T_G$ is the cost to feed one terminal to the LL(1) parser and $|\Gamma|$ is the number of dangling symbols in the derivation tree, which tends to be proportional to the nesting depth of the output code.
In practice, $|\Sigma_T|$ is not so large; JSON has about 15 terminals and \cite{SynCode} reports that Python has 94.
Calculation of $\delta_{a+b}(q_{a+b0}, r.t)$ can be accelerated by precomputing the mapping $\delta_{a+b}^*(q, t)$ for each terminal, DFA state, and LLM token.
At runtime, we calculate the state $q'=\delta_{a+b}(q_{a+b0}, r)$ and lookup the precomputed state $\delta_{a+b}^*(q', t)$ for each terminal sequence $(a,b)$ and token $t$.
This lookup operation can be parallelized into a vector computation across the entire $\mathcal{V}$.
Mask generation is processed by at most $|\Sigma_T|^2$ times of element-wise Boolean and arithmetic operations on the vector of length $|\mathcal{V}|$, which also can be parallelized.
Notice that this cost is usually smaller than the brute force method that searches the shortest terminal sequence by simulating the parser;
The cost is $O(|\Sigma_T|^DT_G)$, where $D$ is the minimum number of terminals in continuation, and $D$ tends to be proportional to the nesting depth of generated sentences.

\subsubsection{Precomputation Phase}
\label{sec:proposed:precomp}
In this phase, we precompute the necessary values required for efficiently calculating Equation~\ref{eqn:fullcost}.
First we calculate $C_a[q]$ provided in Equation~\ref{eqn:c} for each terminal $a \in \Sigma_T$ and $C_{a+b}[q]$ for each two-length terminals $(a,b)$.
To compute them, we use Dijkstra’s algorithm, treating DFA states as nodes, transitions as edges, and token lengths as edge costs.
Next, we estimate $D[A]$ provided in Equation~\ref{eqn:d}.
The computation of $D[A]$ is also based on Dijkstra’s algorithm, where possible derivation states are treated as nodes and derivation steps as edges.
Finally, we precompute the mapping $\delta_{a+b}^*(q, t)$ for each terminal, DFA state, and LLM token.
This is used to efficiently retrieve the DFA state in consuming a remainder and a LLM token illustrated in Figure~\ref{fig:sketch-a}.

\textbf{Space Complexity Analysis.}
The amount of memory for precomputation is the sum of the memory $O(|\Sigma_T|^2|Q|)$ for $C_a[q]$, $O(|\Nonterm|)$ for $D[A]$, and $O(|\Sigma_T|^2|\mathcal{V}||Q|)$ for precomputing mapping $\delta_{a+b}^*(q, t)$, where $|Q|$ is the average size of the DFA states.
Note that the mapping $\delta_{a+b}^*(q, t)$ is sparse because most tokens lead DFAs to a dead state. 

\subsection{Combining TruncProof with Decoding Strategies} \label{sec:strategy} \label{sec:proposed:decode}
TruncProof can be seamlessly integrated with various decoding strategies.
In this work we consider the following three decoding methods:
(1) \textbf{Greedy decoding (Greedy)} is the default strategy in most text-generation libraries.
It takes the token with the best likelihood $P(t\mid t_{<i})$ in each iteration of the text generation.
(2) \textbf{Beam Search (BS)} maintains $b$ best candidates in each iteration and re-selects the $b$ best sequences among the possible continuations.
\cite{picard} adopts BS with their constraint method to improve the accuracy of the generation.
Although BS takes diverse candidates into account and obtains better contents than the greedy strategy, 
it remains difficult to completely avoid future token shortages.
(3) \textbf{Monte Carlo Tree Search (MCTS)} is known to be effective for this type of issue where the selections in beginning have a large effect but their precise value is evaluated in the ending phase.
MCTS originally aims to find the best move in two-person games (\cite{mcts}), but there are some studies for LLM-based text generation (\cite{translation-beyond-beam, ppl-mcts, smc}).
In each generation step $i$, MCTS constructs the search tree whose nodes are possible continuations $t_{<i+k}$ and edges are the selectable next tokens.
MCTS repeats the following stages to grow the search tree: Selection, Expansion, Simulation, and Backup.
In Selection, we traverse the tree up to a leaf based on the following evaluation function introduced by \cite{MasteringGo} that utilizes the likelihood of sequences as a prior:
\begin{equation}
    F(t_{<i}, t) \defeq Q(t_{<i}, t) + c_{puct} P'_\tau(t \mid t_{<i}) \frac{\sqrt{\sum_u N(t_{<i}, u)}}{1 + N(t_{<i}, t)},
\end{equation}
where $Q(t_{<i}, t)$ is the maximum value observed among the continuations of $t_{<i}.t$, $P'_\tau$ is the likelihood modified by the constraint mask and normalized by softmax with temperature $\tau$, $N(t_{<i}, t)$ is the number of investigations beyond $t_{<i}.t$, and $c_{puct}$ is the hyperparameter that balances exploration and exploitation.
In Expansion, we expand the tree to investigate more deeply beyond the leaf which we arrived at.
In Simulation, we apply greedy decoding from the leaf until the end of generation and evaluate the value of the result text $v(t_{<n})$ as the geometric mean of the unmodified likelihood provided directly by the LLM,
which is known as the inverse of the perplexity.
In Backup, we tell the evaluated value $v$ to the ancestors and update their observed values $Q(t_{<i}, t)$.
After some repetitions, we decide the next token $t$ with highest $Q(t_{<i}, t)$.

\section{Experiments and Discussion}
\label{sec:eval}
\subsection{Experimental Setting}
To evaluate TruncProof, we conduct experiments on the JSON-Mode-Eval dataset \cite{NousResearch}, which comprises 100 text-to-JSON tasks.
An example of its prompt is described below:

\begin{lstlisting}[
frame=single,
breaklines=true,
basicstyle={\ttfamily \scriptsize}]
<bos><start_of_turn>user
You are a helpful assistant that answers in JSON. Here's the json schema you must adhere to:
<schema>
{'title': 'WirelessAccessPoint', 'type': 'object', 'properties': {'ssid': {'title': 'SSID', 'type': 'string'}, 'securityProtocol': {'title': 'SecurityProtocol', 'type': 'string'}, 'bandwidth': {'title': 'Bandwidth', 'type': 'string'}}, 'required': ['ssid', 'securityProtocol', 'bandwidth']}
</schema>
I'm currently configuring a wireless access point for our office network and I need to generate a JSON object that accurately represents its settings. The access point's SSID should be 'OfficeNetSecure', it uses WPA2-Enterprise as its security protocol, and it's capable of a bandwidth of up to 1300 Mbps on the 5 GHz band. This JSON object will be used to document our network configurations and to automate the setup process for additional access points in the future. Please provide a JSON object that includes these details.<end_of_turn>
\end{lstlisting}

In this instruction-following task, the goal is to generate syntactically and semantically valid JSON outputs given a natural language prompt.
In \cite{SynCode}, the maximum token limit is fixed at 400, which is approximately six times the average length of the ground truth.
To assess performance under stricter constraints, we define a more challenging configuration, where the maximum token length is dynamically set to $\lfloor L_i^{\text{GT}} \times e \rfloor$ for each instance $i$, with $L_i^{\text{GT}}$ denoting the token length of the ground truth and $e$ an expansion ratio.
Unless otherwise specified, we set $e=1.1$ when comparing TruncProof with other methods.
For completeness, we conduct experiments under various values of $e$. 
\par
As evaluation metrics, we use the following: (1) the percentage of outputs that are grammatically correct, denoted as {\it Syntax}; (2) the percentage of outputs that adhere to the schema specified in the prompt, referred to as {\it Schema}; and (3) the percentage of outputs that are parsed into JSON objects identical to the ground truth, termed {\it Exact-match}.
The last Exact-match metric is newly introduced in this work to specifically assess the semantic validity of the generated JSON outputs.
\par
Notice that the JSON grammar used in \cite{SynCode} does not fully comply with the official JSON standard, RFC 8259.
To ensure a practical and standards-compliant evaluation, we apply an RFC 8259-compliant JSON grammar to all constraint methods when assessing their performance.


\begin{table*}
\caption{Accuracy and generation speed of JSON-mode-eval with $e = 1.1$. 
    Time (ms) denotes the time of generating one token, and the value in parenthesis denotes the overhead of constrained generation, which is calculated by comparing with ``No constraint".
    \dag XGrammar uses its builtin JSON grammar because its format (EBNF) is incompatible with others (Lark).}
    \label{json-accuracy-hard}
    \centering 
    \begin{tabular}{llrrrrr rrrrr}
        \toprule
        & & \multicolumn{5}{c}{Gemma2-2B} & \multicolumn{5}{c}{Llama2-7B-Chat-HF} \\
        \cmidrule(lr){3-7} \cmidrule(lr){8-12}
        & & \multicolumn{3}{c}{Accuracy (\%)} & & & \multicolumn{3}{c}{Accuracy (\%)} & & \\
        \cmidrule(lr){3-5} \cmidrule(lr){8-10}
        Method    & Decoding           & Syntax     & Schema   &  Exact-match & \multicolumn{2}{c}{ Time (ms)}  & Syntax     & Schema   &  Exact-match & \multicolumn{2}{c}{ Time (ms)} \\
        \midrule
         No constraint  & Greedy      & 1 & 1 & 0 & 21.8 & & 2 & 2 & 0 & 17.6 & \\
         \multicolumn{1}{r}{\textit{+prompt}} & Greedy & 8 & 8 & 4 & & & 2 & 2 & 0 & & \\
\midrule
         Outlines\cite{Outlines}      & Greedy      & 36 & 33 & 22 & 458.7 &(+436.9) & 18 & 13 & 4 & 72.2 & (+54.6) \\
                & BS          & 4 & 4 & 2 & 4347.8 & (+4326.0) & 10 & 8 & 4 & 598.8 & (+581.2) \\
\midrule
        Outlines\textit{+prompt} & Greedy& 17 & 17 & 8 & & & 19 & 17 & 5 \\
\midrule
         SynCode\cite{SynCode}   & Greedy      & 4 & 3 & 0 & 23.5 & (+1.7) & 11 & 10 & 4 & 18.4 & (+0.8) \\
                & BS          & 1 & 1 & 0 & 54.0 & (+32.2) & 6 & 6 & 4 & 58.7 & (+41.1)  \\
                 & MCTS       & 4 & 4 & 0 & 438.6 & (+416.8) & 8 & 8 & 4 & 183.5 & (+165.9) \\
\midrule
        SynCode\textit{+prompt} & Greedy & 6 & 6 & 1 & &  & 16 & 14 & 5 \\
\midrule
          XGrammar\cite{XGrammar} \dag  & Greedy      & 5 & 5 & 3 & \bf 22.1 & \bf (+0.3) & 11 & 9 & 2 & \bf 18.3 & \bf (+0.7) \\
            & BS          & 1 & 1 & 0 & 34.3 & (+12.5) & 5 & 3 & 2 & 32.5 & (+14.9) \\
            & MCTS        & 5 & 5 & 2 & 293.3 & (+271.5) & 9 & 8 & 3 & 175.1 & (+157.5) \\
\midrule
        XGrammar\textit{+prompt} & Greedy & 8 & 8 & 4 & & & 16 & 13 & 3 \\
\midrule
          Ours & Greedy    &\bf 100&62 &21 & 25.7 & (+3.9) & \bf 100&51 &2 & 19.0 & (+1.4)\\
          & BS        &\bf 100&85 &37 & 60.8 & (+39.0) & \bf 100&67 &29 & 37.0 & (+19.4) \\
          & MCTS      &\bf 100&\bf 86 & \bf 58 & 518.1 & (+496.3)& \bf 100&\bf 70 & \bf 41 & 209.2 &(+191.6) \\
          \bottomrule
    \end{tabular}
\end{table*}

\textbf{Environment.}
We used 1x H200 GPU to produce all the results.
Beam Search (BS) is performed with 10 beams while Monte Carlo Tree Search (MCTS) is performed with the following hyperparameters: $c_{puct}=5, \tau=2$, 20 trials for each generation step.
It took about 5 seconds to precompute the shortest token lengths for all terminals and nonterminals described in \S\ref{sec:proposed:precomp}.

\subsection{Results}
\label{sec:eval:result}
Table~\ref{json-accuracy-hard} 
presents the results of five approaches: the baseline without any GCG method (denoted as No constraint), Outlines~\cite{Outlines}, SynCode~\cite{SynCode}, XGrammar~\cite{XGrammar}, and our proposed method, TruncProof.
For the No constraint baseline, we adopt Greedy decoding.
All constraint methods except Outlines are evaluated with Greedy, BS, and MCTS.
Note that BS and MCTS are implemented by ourselves, as they are not provided by the original authors.
Following prior work~\cite{SynCode}, we use Gemma2-2B~\cite{gemma_2024} and Llama2-7B-Chat-HF~\cite{llama2} as the underlying language models.
\par
\textbf{Syntax Robustness.}
As expected, under this challenging setting, most outputs generated by the baseline methods are grammatically invalid,
with their Syntax accuracies ranging from only 1\% to 36\%.
This failure occurs mainly because LLMs include excessive whitespace in JSON for readability and thereby waste LLM tokens.
In contrast, TruncProof consistently produces grammatically valid outputs across all decoding strategies and backend LLMs, achieving perfect Syntax accuracy {\it i.e.}, 100\%.
These results clearly demonstrate the effectiveness of our approach in maintaining grammatical correctness under strict token constraints.
\par
\textbf{Accuracy of JSON-mode-eval with prompt engineering.}
To compare the shortening effect of prompt engineering with TruncProof's capabilities, we add the prompt \textit{``Only output JSON. Eliminate white spaces and keep the output as compact as possible."} to the original prompt provided by JSON-Mode-Eval.
Results are shown as \textit{+prompt} in Table \ref{json-accuracy-hard}.
This additional prompt improves the performance slightly in several settings.
As a side effect, unnecessary text such as \verb|```json| is less frequent, leading to a certain degree of gains in the absence of grammar constraints (``No Constraint" rows).
However, it was challenging to ensure LLMs adhere to the maximum token limit when relying solely on prompts.
\par
\begin{figure*}[h]
    \centering
    \includegraphics[width=0.7\linewidth]{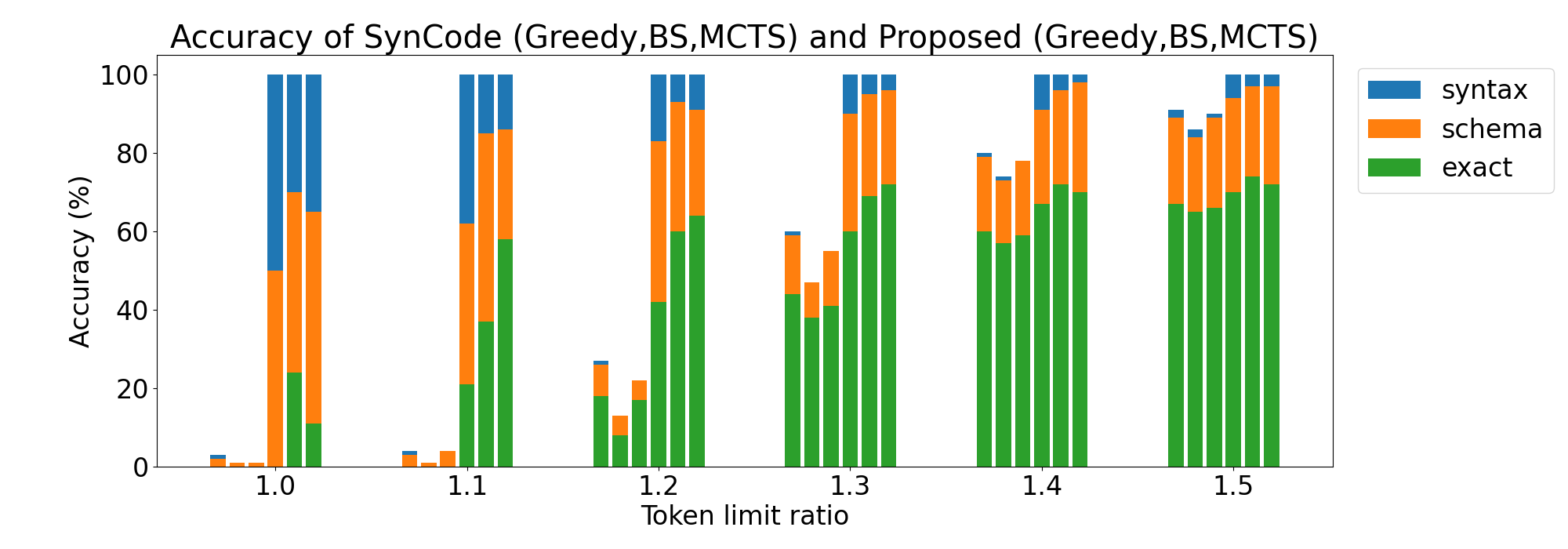}
    \caption{Accuracy of Gemma2-2B with respect to the expansion ratio $e\in[1.0, 1.5]$.
    Six bars drawn in each ratio are the results of SynCode with Greedy decoding, SynCode with Beam Search, SynCode with Monte Carlo Tree Search, ours with Greedy decoding, ours with Beam Search and ours with Monte Carlo Tree Search.
    }
    \label{fig:limits}
\end{figure*}

\textbf{Ranging expansion ratios.}
Figure~\ref{fig:limits} presents the results with different expansion ratios, {\it i.e.}, $e \in [1.0, 1.5]$. 
We observe that our method consistently adheres to the instructed schema, even under strict maximum token limits. 
Moreover, when combined with BS or MCTS, our approach preserves the correctness of the generated content across various expansion settings.
These results experimentally validate the effectiveness of TruncProof in generating grammatically correct outputs, as well as its compatibility with various decoding strategies, which leads to improved semantic quality of the generated texts.

\begin{figure*}[h]
    \centering
    \begin{minipage}[b]{0.8\columnwidth}
    \centering
    \includegraphics[width=\columnwidth]{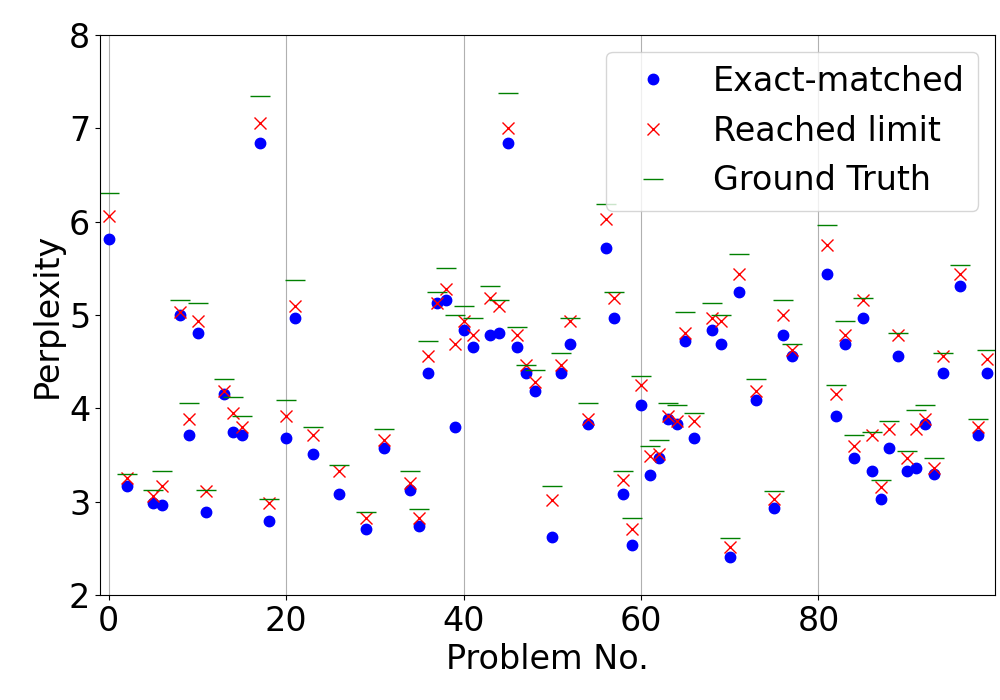}
    \subcaption{SynCode \cite{SynCode}}
    \label{fig:ppl-a}
    \end{minipage}
    \begin{minipage}[b]{0.8\columnwidth}
    \centering
    \includegraphics[width=\columnwidth]{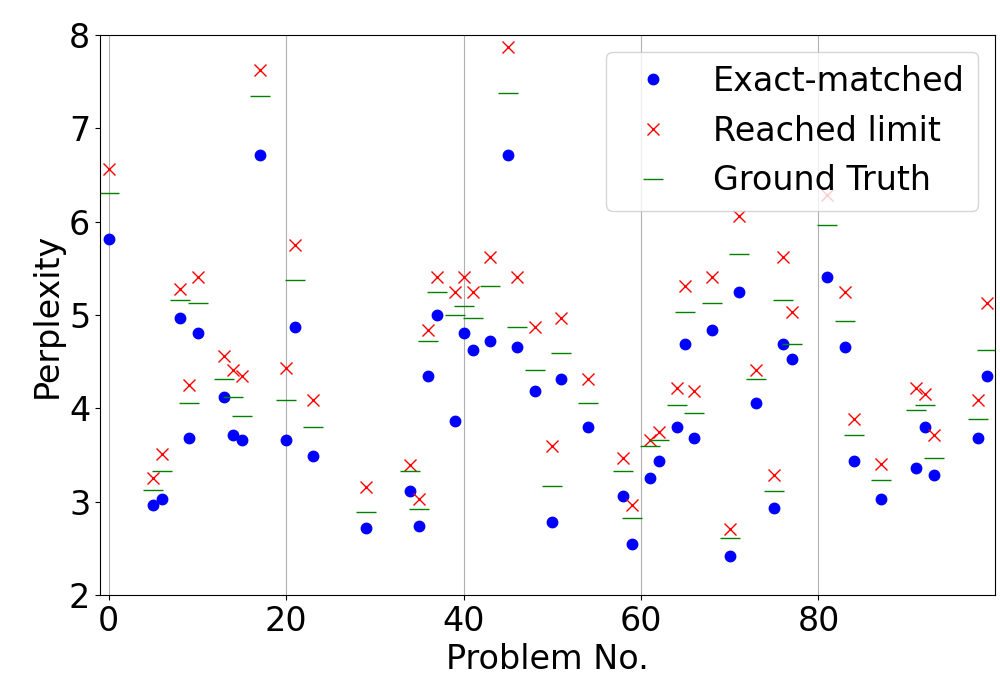}
    \subcaption{TruncProof (Ours)}
    \label{fig:ppl-b}
    \end{minipage}
    \caption{
    The perplexities provided by Gemma2-2B on JSON-Mode-Eval.
    Exact-matched indicates the output whose keys and values are correct under the relaxed token limit. Reached limit indicates the output reached the token limit.}
    \label{fig:ppl}
    \vspace{-0.2cm}
\end{figure*}

\par
\textbf{Semantics Robustness.}
Table~\ref{json-accuracy-hard} 
also shows that when using simple decoding strategies such as Greedy, the Exact-match accuracies of TruncProof remain relatively low (2\%–21\%) although about half (51\%-62\%) of the cases are faithful to the schema.
We emphasize that this outcome is expected; TruncProof only cares about the grammar and the number of tokens, but it does not fully account for the semantic correctness of its outputs.
Also as shown in the same table,
these scores improve significantly when more advanced decoding strategies are employed.
In particular, using BS raises the Exact-match accuracies to 29\%–37\%, and further improvements are observed with MCTS, reaching 41\%–58\%, all while preserving perfect grammatical correctness.
These results highlight the compatibility of TruncProof with various decoding strategies and its ability to enhance semantic quality without compromising syntactic validity.
\par
Also note that such compatibility with various decoding strategies is not necessarily supported by existing methods;
As shown in Table~\ref{json-accuracy-hard}, prior works with BS performs worse than Greedy.
This may be attributed to the presence of many high-likelihood candidates that are grammatically invalid.
To validate this hypothesis, 
in Figure~\ref{fig:ppl}, we visualize
the perplexity of outputs under token shortage (labeled ``Reached limit") for both SynCode \cite{SynCode} and our TruncProof.
As shown, when generation is constrained by SynCode, 
the perplexity of truncated outputs is worse than that of exact-match outputs ({\it i.e.}, successful generations), yet still better than the perplexity of the ground truth
 (see Figure~\ref{fig:ppl-a}). 
This indicates that simply optimizing for likelihood under SynCode may lead to grammatically incorrect outputs due to local optima. 
In contrast, when our method reaches the token limit and generates unnatural outputs, 
the perplexity becomes worse than that of the ground truth, suggesting that TruncProof avoids such invalid local optima by preserving grammatical correctness throughout generation
(see Figure~\ref{fig:ppl-b}).


\section{Limitations}
\label{sec:limit}
As demonstrated in \S~\ref{sec:eval:result}, TruncProof is capable of generating both syntactically and semantically valid JSONs under strict token budget constraints, particularly when paired with advanced decoding strategies.
However, these strategies can slow down the generation process ({\it e.g.}, BS is 2.0-2.4x slower and MCTS is 11.0-20.2x slower than Greedy).
Although successful integration with the strategies is unattainable by other methods, the associated overheads may pose a practical limitation, especially in latency-critical applications.
\par
Furthermore, although this issue is common across GCG methods, enforcing grammatical constraints often distorts the probability distribution produced by the LLM, making it difficult to sample text in a manner that faithfully reflects the model's original conditional probabilities under grammatical correctness. 
To address this, it is important to explore compatibility with methods that approximate the conditional distribution of LLMs under constraints, like \cite{asap}.

\section{Conclusion}
\label{sec:conc}
In this paper, we proposed TruncProof, a novel guardrail to enable LLMs to produce grammatically valid JSONs while adhering to a maximum token limit.
Experiments on the Text-to-JSON instruction tasks \cite{NousResearch} demonstrated that TruncProof can successfully generate syntactically correct outputs even under strict token constraints. 
We also show that TruncProof can be effectively combined with advanced decoding strategies, resulting in outputs that are not only grammatically valid but also semantically accurate. 
Note that TruncProof can potentially be applied to any LL(1) parser ({\it e.g.}, a subset of C provided by \cite{llamacpp}).
For future work, we plan to extend TruncProof to other grammars.
In addition, we will investigate how to accelerate generation when using complex decoding strategies.

\bibliography{citation}
\bibliographystyle{IEEEtran}

\end{document}